\def\year{2019}\relax
\documentclass[letterpaper]{article} 
\usepackage{aaai19}  
\usepackage{times}  
\usepackage{helvet}  
\usepackage{courier}  
\usepackage{url}  
\usepackage{graphicx}  
\usepackage{amsmath,amssymb}
\usepackage{latexsym}
\usepackage{xcolor}
\usepackage[caption=false]{subfig}
\usepackage{multirow}
\usepackage{subfig}
\usepackage[ruled]{algorithm2e}
\usepackage{makecell}
\usepackage{booktabs}
\usepackage{bm}
\graphicspath{{figures/}}

\DeclareMathOperator{\xb}{\mathbf{x}}
\DeclareMathOperator{\db}{\mathbf{d}}
\DeclareMathOperator{\vb}{\mathbf{v}}
\DeclareMathOperator{\zb}{\mathbf{z}}
\DeclareMathOperator{\Ib}{\mathbf{I}}
\DeclareMathOperator{\xib}{\bm{\xi}}
\DeclareMathOperator{\Kb}{\mathbf{K}}
\DeclareMathOperator{\zetab}{\bm{\zeta}}
\newcommand{\RN}[1]{%
	\textup{\lowercase\expandafter{\it \romannumeral#1}}%
}

\begin{document}
%
\title{Self-Adversarially Learned Bayesian Sampling}
\author{Yang Zhao  \\
	State University of New York at Buffalo \\
	yzhao63@buffalo.edu 
	\And Jianyi Zhang  \\ Fudan University \\
	15300180019@fudan.edu.cn
	\And Changyou Chen	\\ State University of New York at Buffalo \\
	cchangyou@gmail.com
}

\maketitle
\begin{abstract}
Scalable Bayesian sampling is playing an important role in modern machine learning, especially in the fast-developed unsupervised-(deep)-learning models. While tremendous progresses have been achieved via scalable Bayesian sampling such as stochastic gradient MCMC (SG-MCMC) and Stein variational gradient descent (SVGD), the generated samples are typically highly correlated. Moreover, their sample-generation processes are often criticized to be inefficient. In this paper, we propose a novel self-adversarial learning framework that automatically learns a conditional generator to mimic the behavior of a Markov kernel (transition kernel). High-quality samples can be efficiently generated by direct forward passes though a learned generator. Most importantly, the learning process adopts a self-learning paradigm, requiring no information on existing Markov kernels, {\it e.g.}, knowledge of how to draw samples from them. Specifically, our framework learns to use current samples, either from the generator or pre-provided training data, to update the generator such that the generated samples progressively approach a target distribution, thus it is called self-learning. Experiments on both synthetic and real datasets verify advantages of our framework, outperforming related methods in terms of both sampling efficiency and sample quality.
\end{abstract}

\section{Introduction}
With the abundance of unlabeled data, Bayesian methods have been increasingly popular in modern machine learning. Various real-world applications have greatly benefited from Bayesian modeling through uncertainty modeling \cite{BlundellCKW:icml15,ZhangLCC:AISTATS18}, deep generative models \cite{FengWL:UAI17,ChenLCWPC:arxiv17} and deep reinforcement learning \cite{osband2016posterior,HaarnojaTAL:ICML17,liu2017stein}. The core of Bayesian methods is efficient Bayesian inference, among which Bayesian sampling stands as one of the most effective tools.

In the setting of big data, recent research has facilitated the development of scalable Bayesian sampling methods. There are mainly two directions on developing these methods, Markov-chain (MC) based and particle-optimization (PO) based methods. Stochastic gradient Markov chain Monte Carlo (SG-MCMC) is a family of scalable MC-based Bayesian learning algorithms designed to efficiently sample from a target (posterior) distribution \cite{WellingT:ICML11,ChenFG:ICML14,DingFBCSN:NIPS14,ChenDC:NIPS15}. Specifically, SG-MCMC generates samples from a Markov chain induced by an It\'{o} diffusion. Under a standard setting, samples from SG-MCMC can approximate a target distribution arbitrarily well given sufficient samples \cite{TehTV:arxiv14,ChenDC:NIPS15}. By contrast, PO-based sampling methods such as Stein variational gradient descent (SVGD) \cite{liu2016stein} initiate a set of particles (or samples) from some simple distributions, and update them iteratively and interactively to approximate a target distribution. Recently, \cite{ChenZWLC:UAI18} proposed a unified Bayesian sampling framework by combing SG-MCMC and SVGD from a Wasserstein-gradient-flow (WGF) perspective, obtaining improved performance compared to both SG-MCMC and SVGD. Our proposed method is partly based on the WGF theory presented in \cite{ChenZWLC:UAI18}.

Though achieving encouraging results, we note two issues in the aforementioned sampling methods: $\RN{1})$ Slow sample generation: though SG-MCMC and SVGD achieve scalable sampling by adopting stochastic gradient information, sample generation is still not efficient enough under complicated models such as a very deep neural network. The problem is even more severe in SVGD as each particle needs to interact with all other particles in the sample-generation process; $\RN{2})$ Slow mixing: samples tend to be highly correlated, leading to slow mixing. Actually, it has been shown that diffusion-based methods such as SG-MCMC might need exponential time to jump out of local modes \cite{RaginskyRT:COLT17,ZhangLC:COLT17}. Thus more sample-efficient algorithms are desperate to be designed.

In this paper, we reinterpret Bayesian sampling as learning a Markov kernel (or a transition kernel), a conditional probability sequentially mapping an old state (sample) to a new state. Leveraging advantages of scalable sampling and recent developments on deep generative models, we reformulate the sampling process based on the generative-adversarial-net (GAN) framework \cite{GoodfellowAMXWOCB:NIPS14}. The formulation is based on the connection between density evolution in a Bayesian sampling algorithm and WGFs. Specifically, a conditional generator which solves the corresponding WGF is trained to mimic the sample-generation process. In this way, both fast sample generation and fast sample mixing are achieved.

We consider two settings in our framework to learn the conditional generators, {\it i.e.}, $\RN{1})$ when samples from the unknown target distribution are available, and $\RN{2})$ when only the form of the target distribution is provided. The former case can be learned by directly adopting standard GAN training techniques, whereas the later case is much more challenging. To overcome the challenge, we propose a {\em self-learning} paradigm that adjust samples from the generator itself to approach the target distribution in a principled way, such that the adjusted samples can be used as real samples to train the generator. We call our proposed framework self-adversarially learned Bayesian sampling. Extensive experiments are performed on both synthetic and real datasets, demonstrating the effectiveness and efficiency of the proposed framework, relative to existing methods.

\section{Preliminaries}
This section reviews background of related Bayesian sampling algorithms, {\it e.g.}, SG-MCMC, SVGD and particle-optimization Bayesian sampling (POS) \cite{ChenZWLC:UAI18}.

\subsection{Stochastic gradient MCMC}
Given observations $\mathcal{D}=\{{\bf d}^{(i)}\}_{i=1}^N$, we aim at drawing samples from a target posterior distribution $p({\bf x}|\mathcal{D})$ with model parameters $\xb\in\mathbb{R}^d$. In Bayesian modeling, we write $p({\bf x}|D)\propto\exp(-U({\bf x}))$, where $U({\bf x}) \triangleq -\log(p(\xb)) - \sum_{i=1}^N{\log p(\db^{(i)}|\xb)}$ is called the potential energy (negative log unnormalized posterior). SG-MCMC is a scalable Bayesian sampling method, which takes stochastic gradient information of the potential energy into consideration. Let $\nabla_{\xb} \widetilde{U}(\xb) \triangleq -\nabla_{\xb}[(\log(p(\xb)) +\frac{N}{n}\sum_{i=1}^n\log p(\db^{\pi_i}|\xb)]$ be a stochastic version of $\nabla_{\xb}U(\xb)$ with $\pi_i$ the $i$-th element of a random permutation of $[1, \cdots, N]$. The stochastic gradient Langevin dynamic (SGLD) stands as the first SG-MCMC algorithm \cite{WellingT:ICML11}, endowing the following update rule (samples are indexed by $\ell$): 
\begin{align}\label{eq:sgld}
\xb_{\ell+1} = \xb_{\ell} -\epsilon_{\ell+1} \nabla_{\xb} \widetilde{U}(\xb_{\ell}) + \sqrt{2\epsilon_{\ell+1}}\zetab_{\ell+1},
\end{align}
where $\{\epsilon_{\ell}\}$ is a stepsize sequence, and $\zetab_{\ell} \sim \mathcal{N}(\mathbf{0}, \mathbf{I}_d)$.
Further development on SG-MCMC methods leads to several variants of SGLD by introducing auxiliary variables into the corresponding dynamics systems \cite{DingFBCSN:NIPS14,ChenFG:ICML14}. With samples $\{\xb_{\ell}\}_{i=1}^M$ from a sampler, one can approximate statistics of a function $f(\xb)$, {\it e.g.}, the posterior expectation is approximated as $\mathbb{E}_{\xb\thicksim p(\xb|\mathcal{D})}[f(\xb)] \approx \frac{1}{M}\sum_{i=1}^M f(\xb_{\ell})$.

\subsection{Stein variational gradient descent}
Different from SG-MCMC, SVGD is derived from a particle-optimization perspective \cite{liu2016stein}. It iteratively and interactively updates a set of particles $\{\xb^{(i)}_{\ell}\}_{i=1}^M$ drawn from some initial distribution. The updating rule follows $\xb^{(i)}_{\ell+1} = \xb^{(i)}_{\ell} + \frac{\epsilon}{M}\Delta\xb^{(i)}_{\ell}$ with 
\begin{align}\label{svgd}
	\begin{split}
	\Delta \xb^{(i)}_{\ell} \triangleq \displaystyle\sum_{j=1}^{M} [-\nabla_{\xb^{(j)}_{\ell}} U(\xb^{(j)}_{\ell})\kappa(\xb^{(i)}_{\ell}, \xb^{(j)}_{\ell}) + \\
	\nabla_{\xb^{(j)}_{\ell}}\kappa(\xb^{(i)}_{\ell}, \xb^{(j)}_{\ell})],
	\end{split}						
\end{align}
where $\kappa(\cdot, \cdot)$ is a positive definite kernel, {\it e.g.}, the RBF kernel $\kappa(\xb^{(i)}, \xb^{(j)}) = \exp(-\|\xb^{(i)}-\xb^{(j)}\|_2^2/h)$ with bandwidth $h$, and $\epsilon$ is the step size.
It is shown that \eqref{svgd} is equivalent to minimizing the Kullback-Leibler (KL) divergence $\textsf{KL}(q(\xb)||p(\xb|\mathcal{D}))$, where $q$ is the underlying density of the particles. Consequently, SVGD drives the particles to asymptotically distributed as the target distribution.

\subsection{Particle-optimization sampling}
Compared with SVGD, an instance of the POS framework \cite{ChenZWLC:UAI18,ZhangZC:arxiv18}, samples from SG-MCMC are likely highly correlated due to the property of Markovian chains. The POS framework alleviate the issue by interpreting both SG-MCMC and SVGD as WGFs on the space of probability measures $\mathcal{P}(\Omega)$, and proposing a unified particle-optimization framework for efficient Bayesian sampling.

Specifically, the POS framework translates Bayesian sampling to solving a partial differential equation defined on $\mathcal{P}(\Omega)$ with $\Omega \subset \mathbb{R}^d$, defined as:
\begin{align}\label{eq:gf_w}
	\partial_{t} \mu_{t} + \nabla \cdot (\vb_{t} \mu_{t}) = 0~.
\end{align}
Here $\mu_{t}$ is an absolutely continuous curve on $\mathcal{P}(\Omega)$ and $\vb_{t}$ is a vector field describing the direction of sample evolutions. In WGFs, $\vb_{t}$ is related to what is known as energy functional $E(\mu)$, mapping a probability measure $\mu$ to a real value, {\it i.e.} $E:\mathbb{R}^d \rightarrow \mathbb{R}$, via the equation \cite{Ambrosio:book05}: $\vb_{t} = -\nabla \frac{\delta E}{\delta \mu_{t}}(\mu_{t})$, 
where $\frac{\delta E}{\delta \mu_{t}}$ is called the {\em first variation} of $E$ at $\mu_{t}$, with evolved directions constrained on the tangent space of the probability manifold. Consequently, gradient flows on $\mathcal{P}(\Omega)$ can be written as
\begin{align}\label{eq:gf_dis1}
\partial_{t} \mu_{t} = -\nabla \cdot (\vb_{t} \mu_{t}) = \nabla \cdot \left(\mu_{t} \nabla(\frac{\delta E}{\delta \mu_{t}}(\mu_{t}))\right)~.
\end{align}
\paragraph{Solving by discrete gradient flows}
An exact solution to the WGF formula \eqref{eq:gf_dis1} is generally infeasible. A typical solution is to approximate the continuous-time solution of \eqref{eq:gf_dis1} with discrete-time flows, called discrete gradient flows (DGFs).
Denote $\mathcal{P}_s(\mathbb{R}^d)$ to be the space of probability measures with finite 2nd-order moments, and define the following optimization problem with a step size $\epsilon$:
\begin{align}\label{eq:proximal}
J_h(\mu) &\triangleq \arg\min_{\nu\in \mathcal{P}_s(\mathbb{R}^d)}\left\{\frac{1}{2\epsilon}W_2^2(\mu, \nu) + E(\nu)\right\}~,
\end{align}
where $W_2^2(\mu, \nu)$ denotes the Wasserstein distance between $\mu$ and $\nu$. Here $E(\nu)$ is such that $p \triangleq \arg\min_{\nu}E(\nu)$ corresponds to the target distribution. The idea of DGFs is to approximate the continuous-time solution $\mu_{t}$ from \eqref{eq:gf_dis1} via a composition of a sequence of $T/\epsilon$ discrete solutions $(\tilde{\mu}_{\ell})_{\ell=1}^{T/\epsilon}$ of \eqref{eq:proximal}, {\it i.e.},

\begin{align}\label{eq:discretegf}
\tilde{\mu}_{\ell} \triangleq J_h(\tilde{\mu}_{\ell-1}) = J_h(J_h(\cdots \mu_0)) \approx \mu_{t}~.
\end{align}
The DGF method is the gradient-descent analogy on Euclidean space for $\nu$. One can show that when $\epsilon\rightarrow 0$, the solution from DGFs \eqref{eq:discretegf} converges to the true flow \eqref{eq:gf_dis1} for all $\ell$ \cite{Craig:thesis14}.

\section{Self-Adversarially Learned Bayesian Sampling}

In this section, we develop a GAN-based framework to efficiently solve the DGF problem \eqref{eq:discretegf}, avoiding the computational complexity of the original particle-approximation-based methods \cite{ChenZWLC:UAI18}. Based on this, more powerful and flexible approximations with a self-adversarial learning scheme are developed.

\subsection{Reformulating POS as conditional GANs}
We first specify the functional energy $E(\nu)$ in \eqref{eq:proximal}. For popular sampling methods such as SG-MCMC and SVGD, $E(\nu)$ has been shown to be the KL-divergence between $\nu(\xb)$ and the target distribution $p(\xb|\mathcal{D})$\footnote{Though the metric for SVGD is defined as a variant of the Wasserstein distance called $\mathcal{H}$-Wasserstein distance \cite{Liu:NIPS17}.}. In this case, the DGF method described above becomes the well-known Jordan-Kinderlehrer-Otto scheme (JKO) \cite{JordanKO:MA98}. For convenience, we instead define the functional energy as the Jensen-Shannon divergence (JSD) between $\nu(\xb)$ and $p(\xb|\mathcal{D})$. Note the JSD also endows the convexity property, rendering a unique optimal solution as for the KL-divergence. Consequently, the DGF \eqref{eq:proximal} becomes
\begin{align}\label{eq:proximal_1}
J_h(\mu) &\approx \arg\min_{\nu\in \mathcal{P}_s(\mathbb{R}^d)}\left\{\frac{\alpha}{2\epsilon}W_2^2(\mu, \nu) + \textsf{JSD}(\nu\|p)\right\},~
\end{align}
where $\alpha$ is a tunable parameter.

\begin{figure}[htb]
	\centering
	\includegraphics[width=0.45\linewidth]{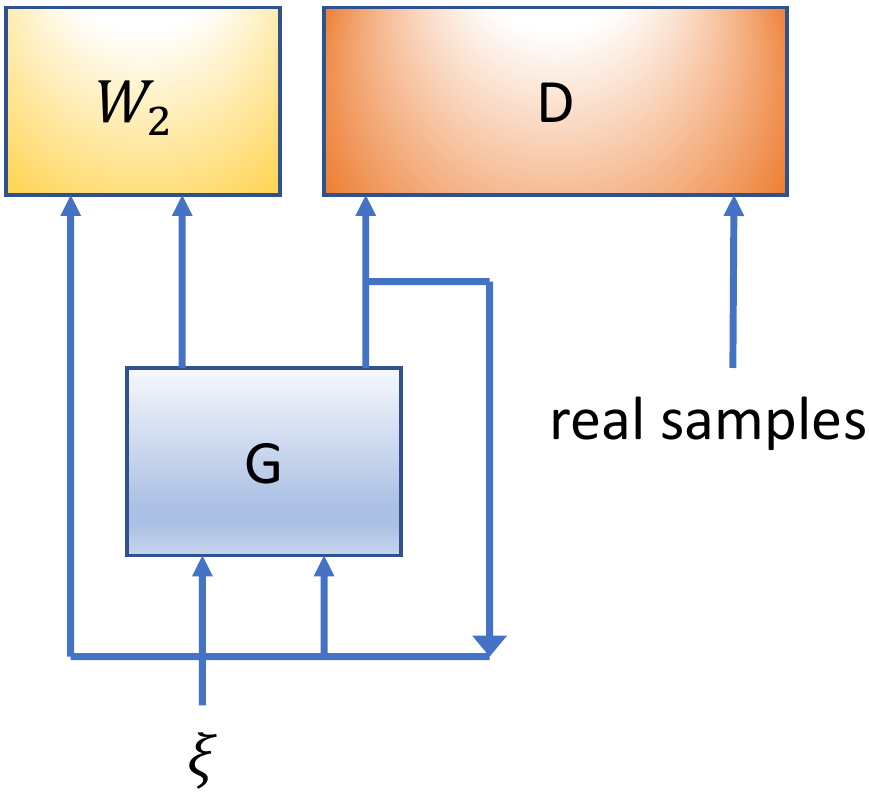}\hspace{0.25cm}
	\includegraphics[width=0.45\linewidth]{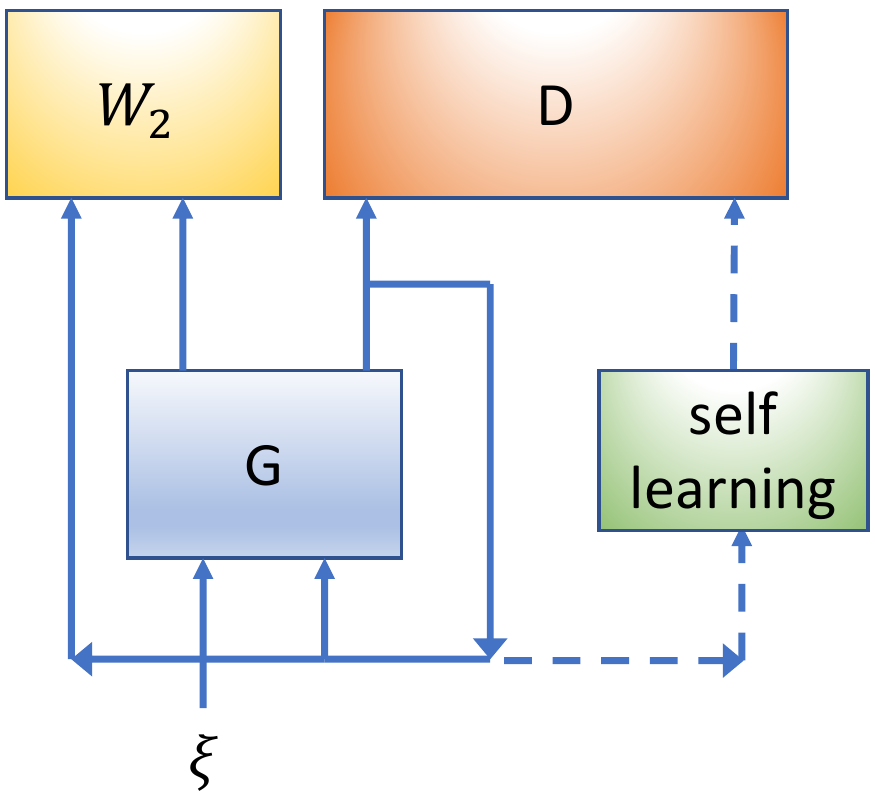}
	\caption{Graphs of the proposed framework with real samples (left) and a self-learning component (right). ``G'' represents generator, ``D'' represents discriminator, and ``$W_2$'' corresponds to the Wasserstein regularizer in \eqref{eq:gan}. Dash lines indicates gradients not flowing back in backpropagation.}\label{fig:fig1}
\end{figure}

Now solving the WGF \eqref{eq:gf_dis1} is equivalent to composing results from a sequence of optimizations defined in \eqref{eq:proximal_1} via \eqref{eq:discretegf}. As a result, $\nu$ is optimized sequentially, each time conditioning on its previous value. In addition, the JSD is well-known to be the objective function of GAN. Consequently, the optimization problem \eqref{eq:proximal_1} can be reformulated as a conditional GAN, where $\nu$ is defined as an implicit distribution induced by a conditional generator $G$. Specifically, $G$ is designed to take an old sample and random noise as input, and outputs the updated sample. The $W_2$ term in \eqref{eq:proximal_1} regularizes the outputs such that they are not too far away from their input samples. According to the GAN theory \cite{GoodfellowAMXWOCB:NIPS14}, \eqref{eq:proximal_1} is equivalent to the following objective:
\begin{align}\label{eq:gan}
	&\mathcal{L} = \min_{G}\max_{D}\mathbb{E}_{\xb \sim p_{d}(\xb)}\left[\log D(\xb)\right] \\
	+& \mathbb{E}_{\zb\sim p_{z}(\zb)}\left[\log(1 - D(G(\tilde{\xb}, \zb)))\right] + \frac{\alpha}{2\epsilon}W_2^2(p_G(\tilde{\xb}), p(\tilde{\xb}))\nonumber
\end{align}
where $p_{d}(\cdot)$ denotes the true data distribution, $\tilde{\xb}$ is the previous sample from the generator $G$ whose implicit distribution is denoted as $p(\tilde{\xb})$; $p_G(\tilde{\xb})$ denotes the implicit distribution of the output $G(\tilde{\xb}, \zb)$; and $D(\cdot)$ is a discriminator network to distinguish an input to be real or fake. A conditional generator is required because the output is correlated with the input via the $W_2$ term. The objective \eqref{eq:gan} is illustrated in Figure~\ref{fig:fig1} (left).

We note two differences between our formulation \eqref{eq:gan} and the standard GAN: $\RN{1})$ Performing stochastic gradient descend (SGD) for learning is challenging due to the existence of the $W_2$ regularization term; and $\RN{2})$ The real sample $\xb$ might not be available in training. In the following, we address the first problem by deriving an approximate form for $W_2$ in Section~\ref{sec:adv}, by assuming the availability of real/true data $\xb$. We then proceed to solve the second problem by proposing a self-adversarial-learning component to automatically adjust samples from the generator such that they approach real samples in Section~\ref{sec:sadv}.

\subsection{Adversarially-learning Bayesian sampling}\label{sec:adv}
\paragraph{Approximating the Wasserstein regularizer}
Following \cite{ChenZWLC:UAI18}, we use particle approximation to deal with the Wasserstein term in \eqref{eq:gan}. Let $(\tilde{\xb}_{-}^{(i)})_{i=1}^M$ be a minibatch of input samples from the generator (samples from last step), and $(\tilde{\xb}^{(i)})_{i=1}^M$ be the output samples. Using samples/particles to approximate the $W_2$ in \eqref{eq:gan}, $W_2^2(p_G(\tilde{\xb}), p(\tilde{\xb}))$ is approximated as
\begin{align}\label{eq:w2}
	W_2^2&(p_G(\tilde{\xb}), p(\tilde{\xb})) \approx \inf_{p_{ij}}\sum_{i,j}p_{ij}c(\tilde{\xb}^{(i)}, \tilde{\xb}_{-}^{(j)}) \\
	s.t.&~~ \sum_j p_{ij} = \frac{1}{M}, ~~\sum_i p_{ij} = \frac{1}{M}~,\nonumber
\end{align}
where $c(\tilde{\xb}^{(i)}, \tilde{\xb}_{-}^{(j)}) \triangleq \|\tilde{\xb}^{(i)} - \tilde{\xb}_{-}^{(j)}\|_2^2 \triangleq c_{ij}$. Now the goal is turned into solving for the optimal $\{p_{ij}\}$. Introducing Lagrangian multipliers $\{\alpha_i, \beta_i\}$ to deal with the constraints, and adding an entropy regularized term for $\{p_{ij}\}$, the dual problem can be written as
\begin{align}\label{eq:dual}
	\begin{split}
		\mathcal{L}&(\{p_{ij}\}, \{\alpha_i\}, \{\beta_i\}) = \lambda\sum_{i,j}p_{ij}\log p_{ij} + p_{ij}c_{ij} \\
		&+ \sum_i \alpha_i(\sum_j p_{ij} - \frac{1}{M}) + \sum_j \beta_j(\sum_i p_{ij} - \frac{1}{M})
	\end{split}
\end{align}
Solving for \eqref{eq:dual}, the optimal $p_{ij}$ endows a forms of $p_{ij}^* = u_i e^{-c_{ij}/\lambda}v_j$, where $u_i \triangleq e^{-\frac{1}{2}-\frac{\alpha_i}{\lambda}}$, $v_j = e^{-\frac{1}{2}-\frac{\beta_j}{\lambda}}$. Now substituting the optimal $p_{ij}^*$ back to \eqref{eq:w2}, the Wasserstein distance can be approximated as:
\begin{align}\label{eq:W2_app}
	W_2^2&(p_G(\tilde{\xb}), p(\tilde{\xb})) \approx \gamma \sum_{i,j}c_{ij}e^{-c_{ij}/\lambda}~,
\end{align}
where the original $u_i$ and $v_j$ have been merged into the constant $\gamma$ for simplicity. 

\paragraph{Adversarially-learned Bayesian sampling}
Given the approximation \eqref{eq:W2_app} for $W_2$ and real training data $\xb$, gradients of the generator parameters can be readily calculated by backpropagation, making generator update with SGD readily available. We call this version of our framework Adversarially-learned Markovian chain (AL-MC), with detailed algorithm given in the Supplementary Material (SM) on our homepage.

\subsection{Self-adversarially learned Bayesian sampling}\label{sec:sadv}

A more challenging setting in practice is that real samples are not readily available, {\it e.g.}, in posterior sampling where only an (unnormalized) posterior distribution is provided. This section addresses the problem of how to learn a generator to generate effective samples only with such information.

Our basic idea is to add a {\em self-learning} module that can automatically adjust the current samples from the generator to approach a target distribution. These adjusted samples are then used as real data to update both the generator and discriminator gradually. This idea is illustrated in Figure~\ref{fig:fig1} (right). Based on an unnormalized target distribution, we consider two settings:

\paragraph{Real data generation with approximate Bayesian sampling}
In this case, one is assume to be able to directly draw approximate samples from a target distribution, based on the previous outputs of the generator. This procedure can be done by adopting existing effective approximate samplers such as SG-MCMC and SVGD. Specifically, let the previous outputs from the generator be $\{\tilde{\xb}_t^{(i)}\}_{i=1}^{M}$, the real samples for next generator update are then approximated as
\begin{align*}
	\begin{split}
	\{\xb_{t+1}^{(i)}\}_{i=1}^{m} = \; &\textsf{SG-MCMC}(\{\tilde{\xb}_t^{(i)}\}_{i=1}^{m}) \quad \mbox{or}\\
	&\textsf{SVGD}(\{\tilde{\xb}_t^{(i)}\}_{i=1}^{m})~,
	\end{split}~
\end{align*}
where $\textsf{SG-MCMC}(\cdot)$ or $\textsf{SVGD}(\cdot)$ means running one or several SG-MCMC/SVGD updates on the input samples toward the target distribution. Based on these approximate real samples, the generator update then proceeds as what AL-MC does described in the last section.

One can easily see that the effectiveness of this learning scheme highly depends on the approximate sampling algorithms, {\it e.g.}, SG-MCMC or SVGD. Empirically, we usually observe samples from the generator collapsed to one mode on a multi-mode target distribution. The reason is that when modes are too far away from each other, making samples jump from one mode to another with SG-MCMC or SVGD typically takes a long time, and sometimes they even fail to move. This misleads the generator to be trained to generate samples only on one mode of the target distribution. In the following, we propose a novel self-adversarial learning module to overcome this issue, which does not even need samples from an approximate sampler.

\paragraph{Real-data generation with self learning}
Developed on ideas from \cite{HanL:ICML18}, we propose a self-learning scheme which can automatically adjust samples from a generator by only relying on gradient information from the current sample density, instead of on the true gradient information of the target distribution. Specifically, let the current induced distribution from the generator be $\nu$ (an implicit distribution without an explicit form) with corresponding samples $\{\xb^{(1)}, \cdots, \xb^{(M)}\}$. The output of the self-learning module in Figure~\ref{fig:fig1} (right) is defined as:

\begin{align}\label{eq:ag-svgd}
	\xb^{(i)} \leftarrow \xb^{(i)} + \frac{\epsilon}{Z}\sum_{j=1}^M\omega_j&\left[\nabla_{\xb^{(j)}}\log \nu(\xb^{(j)})\kappa(\xb^{(i)}, \xb^{(j)})\right.\nonumber\\
	&\left. + \nabla_{\xb^{(j)}}\kappa(\xb^{(i)}, \xb^{(j)})\right],~
\end{align}
where $\omega_j = \nu(\xb^{(j)}) / p(\xb^{(j)}|\mathcal{D})$, and $Z \triangleq \sum_j\omega_j$. According to \cite{HanL:ICML18}, the distribution of $\{\xb^{(i)}\}$ is guaranteed to asymptotically converge to $p(\xb|\mathcal{D})$. 

Note the issue in the above update is that there are no explicit forms for both $\nu(\xb)$ and $\nabla_{\xb}\log\nu(\xb)$ as $\nu$ is an implicit distribution. Therefore, we adopt density estimation techniques for approximation. For $\nu(\xb)$, we use the popular kernel density estimator \cite{Guidoum:15}, {\it i.e.},

\begin{align}
	\nu(\xb) \approx \frac{1}{M}\sum_{i=1}^M\kappa^*(\xb, \xb^{(i)}),~
\end{align}
where $\kappa^*$ is a positive kernel in our paper. For $\nabla_{\xb}\log\nu(\xb)$, we apply the recently developed Stein gradient estimator \cite{LiT:ICLR18}, defined as

\begin{align}
	\nabla_{\xb}\log\nu(\xb) \approx -\left(\kappa + \eta \Ib\right)^{-1}\sum_{i=1}^M\nabla_{\xb^{(i)}}\kappa^*(\xb, \xb^{(i)})~
\end{align}
where $\kappa$ is a kernel matrix with $\Kb_{ij}=\kappa^*(\xb^{(i)}, \xb^{(j)})$. The Stein gradient estimator has been shown to significantly outperform existing gradient estimator methods \cite{LiT:ICLR18}. Note we use two forms of RBF kernels $\kappa^*(\xb^{(i)}, \xb^{(j)})$ with bandwidth $h^*$ and $\kappa(\xb^{(i)}, \xb^{(j)})$ with bandwidth $h$ to allow flexibility. $\kappa$ is used in SVGD, and $\kappa^*$ is used to approximate the density and the log-posterior gradient.
Posterior sampling with \eqref{eq:ag-svgd} is an instance of the SVGD-without-gradient framework \cite{HanL:ICML18}. We denote the update \eqref{eq:ag-svgd} as approximate gradient SVGD (AG-SVGD), and will show impressive performances compared to standard SG-MCMC or SVGD in the experiments.

Taking AG-SVGD as the self-learning module in Figure~\ref{fig:fig1} (right), we obtain what is called self-adversarially-learning Markov chain (SAL-MC) sampler, described below.

\paragraph{The SAL-MC sampler}
The training procedure is given in the first part of Algorithm \ref{sal_alg}, where AG-SVGD is used to generate approximate real samples from a target distribution. This procedure would gradually guide the generator to generate real samples as the stationary distribution of AG-SVGD is the target distribution. In addition, this self-adjusted behavior makes samples jump out of local modes easier, leading to much better performance compared to the one using SG-MCMC or SVGD to generate real samples, as will be shown in the experiments. We also find that multiple updates of the discriminator per generator update can increase effective sample size (ESS). 
\begin{algorithm}[htb]\label{sal_alg}
	\SetKwInOut{Input}{Input}
	\SetKwInOut{Output}{Output}
	
	\Input{$\tilde{\xb}_0\sim \mathcal{N}(\mathbf{0}, \Ib)$, $\xib \sim \mathcal{N}(\mathbf{0}, \sigma^2\Ib)$, $m$ and $S=\emptyset$}
	\Output{Transition operator {\it G} and samples set $S$}
	1. SAL-MC training\;
	\For{t=$1, 2\cdots $}
	{
		{\it // Adjust particles}:\\
		$\xb_t \leftarrow$ AG-SVGD($\tilde{\xb}_{t-1}$)
		
		{\it // Adversarial training between} $\xb_t$ {\it and} $G(\tilde{\xb}_{t-1}, \xib)$:
		
		\For{i=$1, 2\cdots m$}
		{
			train the discriminator with the objective \eqref{eq:gan}
		}
		train the generator with the objective \eqref{eq:gan}
		
		{\it // Update adversarially-learned particles}:
		
		$\tilde{\xb}_t \leftarrow G(\tilde{\xb}_{t-1},\xib)$
		
		\For{i=$1, 2\cdots m$}
		{
			$\tilde{\xb}_t \leftarrow G(\tilde{\xb}_t,\xib)$
		}
	}
	2. SAL-MC sampling\;
	\For {l=$1, 2\cdots $}
	{
		$\tilde{\xb}_l \leftarrow G(\tilde{\xb}_{l-1}, \xib)$
		
		no MH step and directly add $\tilde{\xb}_l$ to $S$
	}
	\caption{SAL-MC training and sampling}
\end{algorithm}
After training, only the conditional generator $G$ is adopted to generate a sequence of samples via the following generative process:
\begin{align}\label{eq:gen}
	\tilde{\xb}_{t+1} = G(\tilde{\xb}_t, \xib)~,
\end{align}
where $\xib$ is a random noise drawn from a simple distribution such as an isotropy multivariate Gaussian $\mathcal{N}(\mathbf{0}, \sigma^2\Ib)$. In this way, $G$ is taken as a transition operator of a Markov chain.

Unlike SG-MCMC and SVGD, no gradient information from the target distribution is needed, leading to fast sample generation only through forward passes of a neural network. In addition, the generator allows distant jumps between consecutive samples via the complex transformation in $G$, making samples mix faster. In practice, we do not apply any burn-in steps or thinning methods on the samples, while it still shows a good convergence property with high ESS, as will be demonstrated in the experiments.

\section{Experiments}
In this section, we first examine the effectiveness of AG-SVGD with stochastic gradient estimations for direct sampling, and compare it with standard SGLD and the recently proposed Annealed-SVGD (A-SVGD) \cite{HanL:ICML18}. We then apply the proposed SAL-MC framework on a set of multi-mode synthetic distributions, as well as on Bayesian Logistic Regression (BLR) tasks. We compare SAL-MC with the recently developed A-NICE-MC method \cite{Songnice:NIPS2017}, which is also based on adversarial training. Finally, we apply our AL-MC algorithm for image generation trained on real samples.

\subsection{Verification of AG-SVGD}
We conduct experiments on two multi-mode toy examples, a 5D Gaussian Mixture Model (GMM) distributed in an aggregation state and a challenging 2D-GMM distribution with distant modes and varied variances. We use the same RBF kernel and the median trick for AG-SVGD and A-SVGD. As suggested by \cite{WellingT:ICML11}, a polynomially-decayed step size ${\epsilon_{t}=a/(t+1)^{0.55}}$ is used in SGLD for a fair comparison. We use samples to approximate the mean and variance of the distributions, measured by mean squared errors (MSE) w.r.t.\! true values. The results are averaged over 20 random runs with 500 iterations in each run. To be consistent with \cite{HanL:ICML18}, we also use RBF kernels and the median trick when calculating maximum mean discrepancy (MMD) between the sample approximation and true distribution for all the three methods.

\begin{figure}[htb]
	\centering
	\includegraphics[width=1\linewidth]{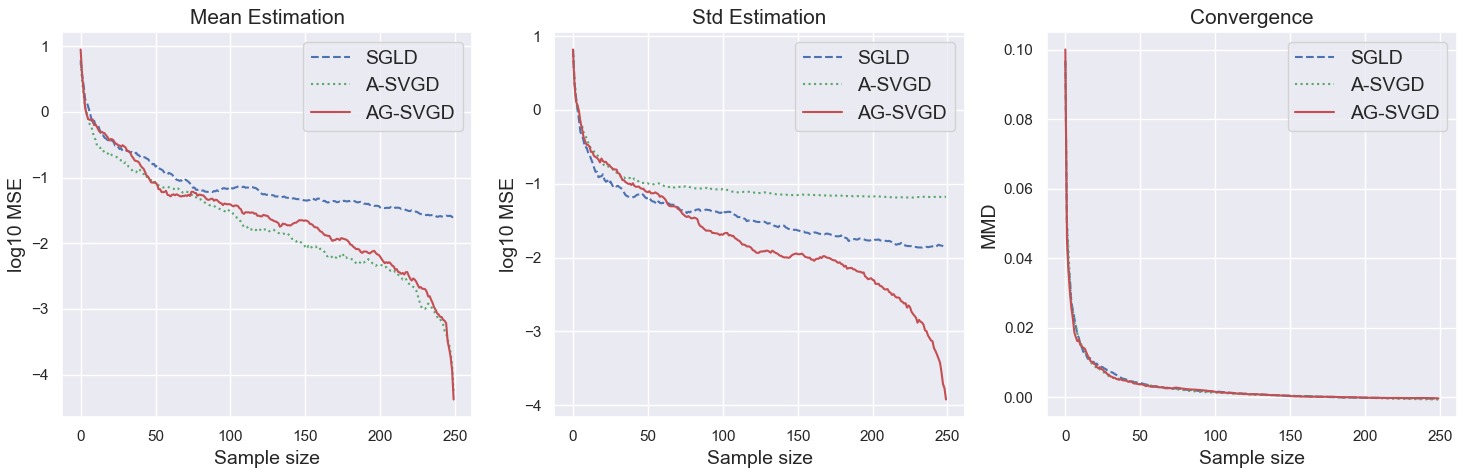}
	\caption{Simulation results on \textit{Mog10}}\label{fig:mog10}
\end{figure}

The relatively simple 5D mixture distribution endows 10 modes. Figure~\ref{fig:mog10} compares the convergence results of the three methods by varying sample sizes. Different from A-SVGD, our AG-SVGD does not need gradient information from the target distribution. Surprisingly, however, AG-SVGD is comparable to A-SVGD when the number of particles is big enough. Furthermore, AG-SVGD obtains the best estimation of variance among the three algorithms. Moreover, AG-SVGD has a slightly better convergence property when sample size is small. In contrast, SGLD performs the worse due to the noisy updates and slow mixing samples. 
\begin{figure}[htb]
	\centering
	\includegraphics[width=1\linewidth]{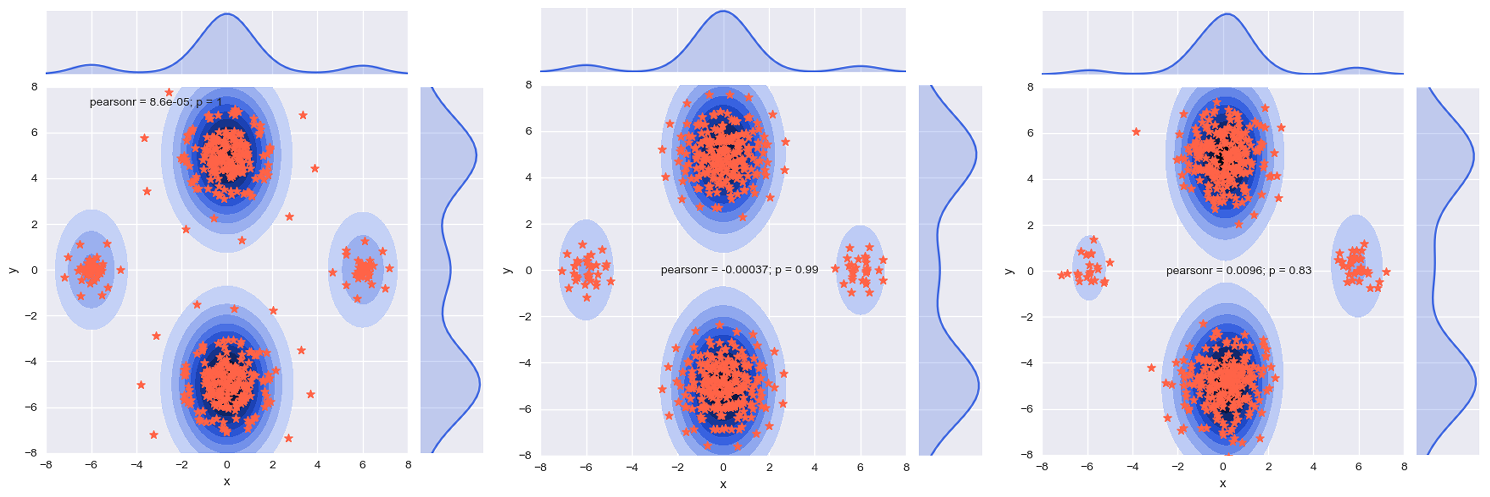}
	\caption{Samples on \textit{Mog4}. From left to right: AG-SVGD, A-SVGD and SGLD. 
	}\label{fig:fig4}
\end{figure}
\begin{figure}[htb]
	\centering
	\includegraphics[width=1\linewidth]{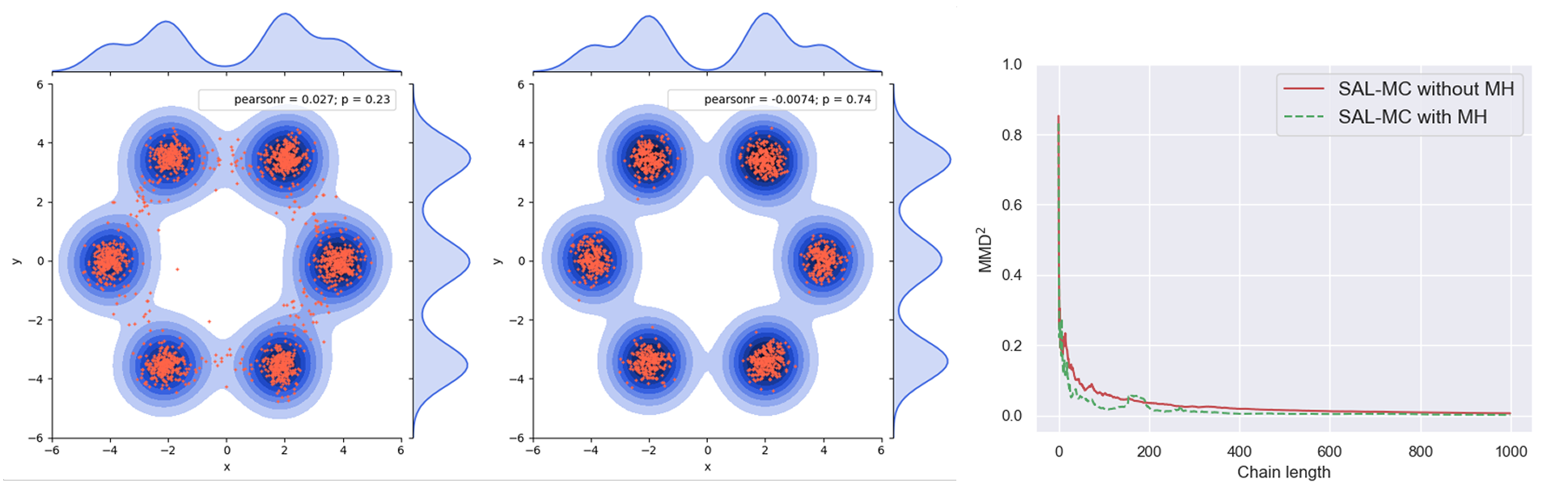}
	\caption{SAL-MC on \textit{Mog6}. Left: without MH; Middle: with MH. }\label{fig:mog6}
\end{figure}
\begin{figure}[htb]
	\centering
	\includegraphics[width=1\linewidth]{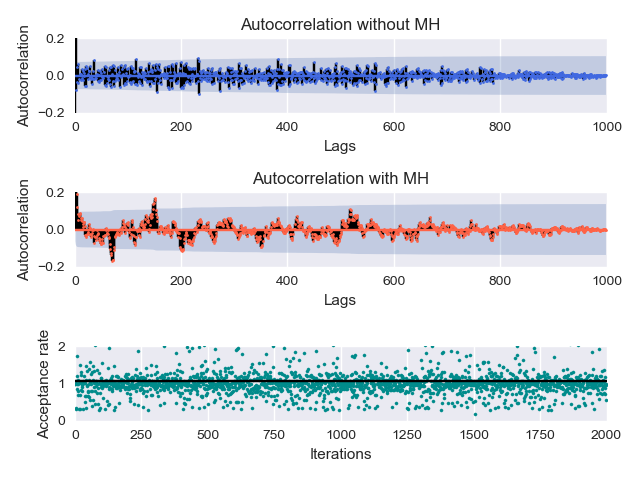}
	\caption{Convergence behavior of SAL-MC on \textit{Mog6}. Acceptance rate in the third plot (without MH steps) denotes the ratio of the proposed-sample probability to current-sample probability, {\it i.e.}, $p(\tilde{\xb}_{t+1})/p(\tilde{\xb}_{t})$)}\label{fig:act}
\end{figure}
\begin{figure}[htb]
	\centering
	\includegraphics[width=1\linewidth]{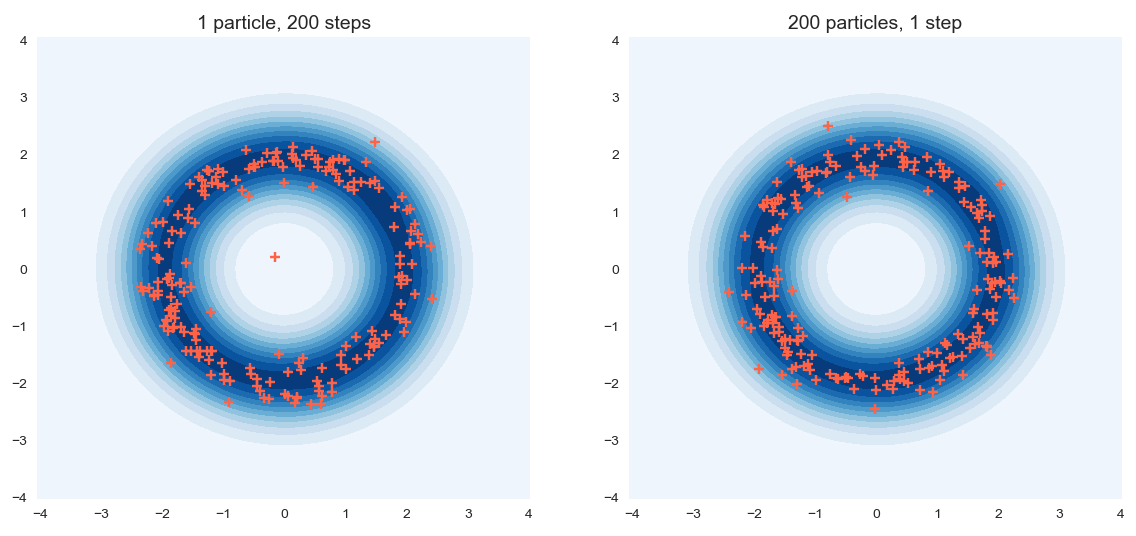}
	\caption{SAL sampler results of \textit{Ring} under two settings}\label{fig:step}
\end{figure}
\begin{figure*}[t]
	\centering
	\includegraphics[width=0.9\linewidth]{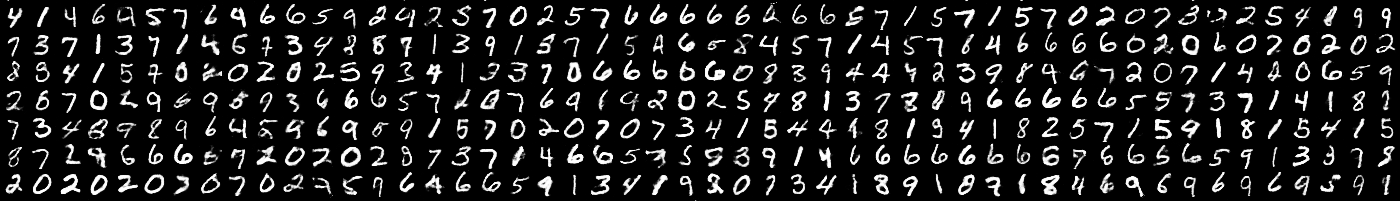}
	\includegraphics[width=0.9\linewidth]{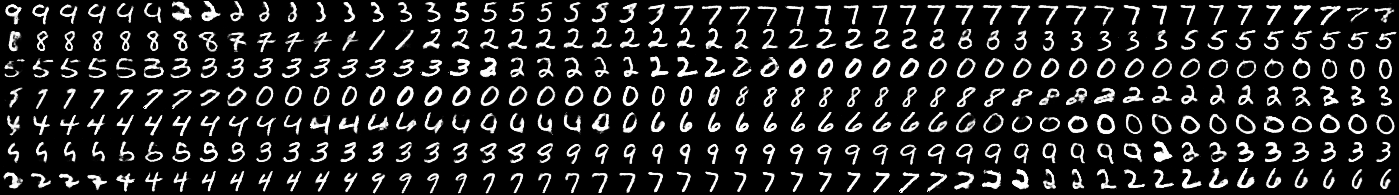}
	\caption{Sampling results of digits for $\xib \sim \mathcal{N}(\mathbf{0}, 0.5\mathbf{I})$ (top) and $\xib=0$ (bottom). Each row represents 50 consecutive samples from the same chain. The sampler with $\xib \sim \mathcal{N}(\mathbf{0}, 0.5\mathbf{I})$ mixes well by generating samples from different modes easily.}
	\label{fig:mnist}
\end{figure*}

For the challenging 2D-GMM dataset {\it Mog4}, the samples obtained are visualized in Figure~\ref{fig:fig4}. Again, we use the same bandwidth of $h=0.01*median$ for AG-SVGD and A-SVGD. Parameters of the gradient estimator are set among $h^* = \left\{0.5, 1, 2, 4, 8, 10\right\}$ and $\eta = \left\{0.05, 0.1, 1, 5, 10, 15\right\}$, selected by a grid search. The initial particles are empirically drawn from $\mathcal{N}(\mathbf{0}, 2.5\mathbf{I})$. As can be seen, both A-SVGD and SGLD somewhat fail to balance samples from different modes, whereas our AG-SVGD is able to travel better between low and high density regions, leading to a more accurate approximation. 
In the following, the well-validated AG-SVGD is applied in SAL-MC samplers as the self-learning module, which is tested on a number of datasets.
\begin{table}[htb]
	\centering
	\caption{ESS (2000 maximum) and ESS speed on synthetic distributions of both methods}
	\resizebox{1\linewidth}{10mm}{
		\begin{tabular}{c|c|c|c|c|c}
			\toprule[1pt]
			\multicolumn{2}{c|}{ESS / (ESS/s)}       & \textit{Ring}       & \textit{Mog2} & \textit{Mog6}       & \textit{Ring5}\\ \midrule[0.7pt]
			\multirow{2}{*}{SAL-MC} & -   & \textbf{1635/121138} & \textbf{1435/72691} & \textbf{1212/65287} &-\\ \cline{2-6} 
			& MH & 1561/25624  & 1172/17531 & 978/12235  &\textbf{414/16287}\\ \hline
			\multirow{2}{*}{A-NICE-MC} & -   & N/A         & N/A     & N/A   & N/A\\ \cline{2-6} 
			& MH & 2000/43887  & 951/17216 & 889/11745   &335/6022\\ 
			\bottomrule[1pt]
		\end{tabular}
	}
	\label{tab:tab1}
\end{table}

\subsection{SAL-MC on synthetic datasets}
Next we demonstrate the effectiveness of SAL-MC by comparing it with A-NICE-MC \cite{Songnice:NIPS2017}. We adopt the \textit{Ring}, \textit{Mog2}, \textit{Mog6} and \textit{Ring5} datasets used in A-NICE-MC, and measure the efficiency of a MCMC method in terms of ESS (2000 maximum) and ESS per second. The smallest ESS among all dimensions is reported. More details are provided in SM. The stochastic term $\xib$ in the algorithm is drawn from $\mathcal{N}(\mathbf{0}, 5 \Ib)$ for all experiments. Since A-NICE-MC requires a Metropolis-Hastings (MH) step to accept or reject a sample, we also test the MH step in our algorithm.

The results are shown in Table \ref{tab:tab1}, where SAL-MC consistently outperforms A-NICE-MC. Interestingly, A-NICE-MC collapses without MH steps, whereas SAL-MC works similarly with or without a MH step. In addition, it is observed that both POS and SGLD achieve a very low ESS that is around 10. We also calculate the Gelman-Rubin convergence statistic $\hat{R}$ \cite{gelman1995bayesian,brooks1998general}, a common convergence diagnostic using multiple chains to check for the convergence of an algorithm. Typically, a values of $\hat{R}$ close to one ({\it e.g.}, less than 1.1) is a good indicator of convergence. It is observed SAL-MC obtains $\hat{R} = 1.00$ using 32 chains for all tasks.

We further illustrate samples drawn from the \textit{Mog6} example in the first two scatter plots of Figure~\ref{fig:mog6}. As can be seen, SAL-MC is able to learn the six modes reasonably well, no matter if it is with or without the MH step. Without MH, SAL-MC tends to be able to generate more samples in low-density regions in between different modes. Note appropriate injected noise $\xib$ should be chosen because too small noise makes distant jumps difficult, while too large noise makes the convergence slower with fewer effective samples. The rightmost plot of Figure~\ref{fig:mog6} together with Figure~\ref{fig:act} demonstrate the proposed SAL-MC endows nice convergence properties in terms of {\em rapid decay of MMD} (efficiency and low bias), {\em low autocorrelation} (low variance), and {\em high acceptance rates}.

We also compare the settings of single and multiple chains in SAL-MC with equal number of samples. Specifically, for the single-chain setting, a sample is initialized from $\xb_0 \sim \mathcal{N}(\mathbf{0}, \mathbf{I})$, followed by 200 updates from SAL-MC to form 200 samples. For the multiple-chain setting, we initialize 200 samples followed by a single-step update to form the final 200 samples. The results on \textit{Ring} are visualized in Figure \ref{fig:step}, from which we can see both settings perform similarly with well approximate samples from the target distribution. 

\begin{figure}[htb]
	\centering
	\includegraphics[width=1\linewidth]{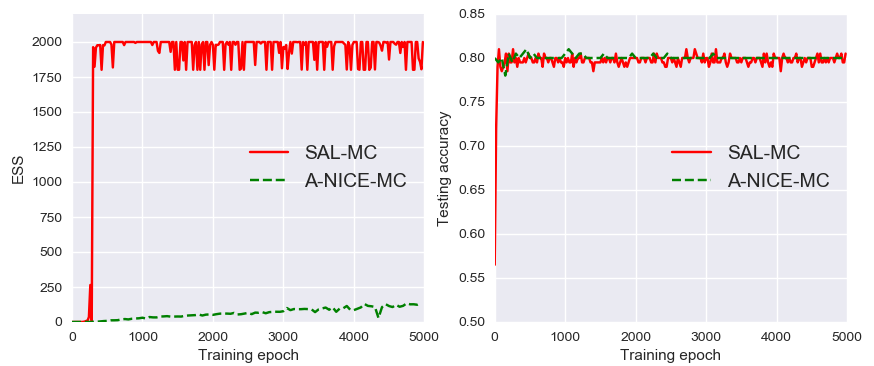}
	\caption{ESS and test accuracy with respect to training epochs averaged by 10 different runs. ({\it German} dataset)}\label{fig:ess}
\end{figure}
\begin{figure}[htb]
	\centering
	\includegraphics[width=0.95\linewidth]{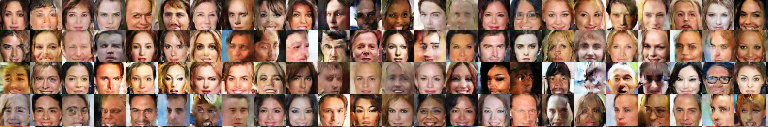}
	\includegraphics[width=0.95\linewidth]{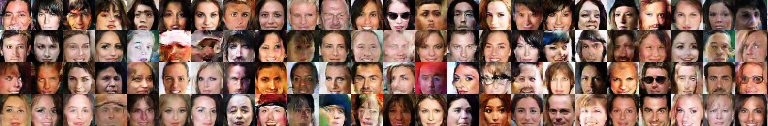}
	\caption{Sampling results of faces for $\xib \sim \mathcal{N}(\mathbf{0}, 0.5\mathbf{I})$ (top) and $\xib=0$ (bottom). Each row represents 24 consecutive samples from the same chain.}
	\label{fig:celeba}
\end{figure}

\subsection{SAL-MC for BLR}
We further compare SAL-MC with A-NICE-MC on several BLR tasks for scalable Bayesian posterior sampling. Three datasets, {\it Heart} (532-13), {\it Australian} (690-14) and {\it German} (1000-24), are used, where $(a$-$b)$ means a dataset with $a$ samples and feature dimensionality $b$. The mini-batch size for training is 64; and the injected noise $\xib$ is drawn from $\mathcal{N}(\mathbf{0}, \Ib)$ for all tasks.

The ESS and ESS speed are shown in Table~\ref{tab:tab2}, which are calculated after 20K training epochs. With a MH step, SAL-MC tends to generate relatively highly-correlated samples with high rejection rates. Nerveless, it is still better than A-NICE-MC, which, unfortunately, does not even work without MH steps. We also evaluate BLR in terms of testing accuracies, which are calculated by averaging over 10 runs with 32 chains. The models are trained on a random 80\% of the datasets and tested on the remaining 20\% in each run. The results are respectively 84.10\%, 88.38\% and 80.32\% on {\it Heart}, {\it Australian} and {\it German} datasets for SAL-MC, which are the same as A-NICE-MC \cite{Songnice:NIPS2017}. However, as shown in Figure \ref{fig:ess}, SAL-MC obtains much better ESS. The experiments also indicate A-NICE-MC must take 20K training iterations to get the highest ESS (which is much lower than SAL-MC as shown in Table~\ref{tab:tab2}).
\begin{table}[htb]
	\centering
	\caption{ESS (2000 maximum) and ESS speed for BLR.}
	\resizebox{1.0\linewidth}{12mm}{
		\begin{tabular}{c|c|c|c|c}
			\toprule[1pt]
			\multicolumn{2}{c|}{ESS / (ESS/s)}       & {\it Heart}        & {\it Australian}     & {\it German}  \\ \midrule[0.7pt]
			\multirow{2}{*}{SAL-MC} & -   & \textbf{1683/93140}  & \textbf{1385/70655} &\textbf{1897/86512}\\ \cline{2-5} 
			& MH & 1/10        & 1/8            & 1/14  \\ \hline
			\multirow{2}{*}{A-NICE-MC} & -   & N/A         & N/A            & N/A  \\ \cline{2-5} 
			& MH & 663/10939    & 596/9834& 483/7848 \\
			\bottomrule[1pt]
	\end{tabular}}
	\label{tab:tab2}
\end{table}

\subsection{AL-MC for image synthesize}
We finally test AL-MC, a variant with only real samples available in training, for image synthesis on MNIST and CelebA datasets. The balance factor of Wasserstein regularization term in \eqref{eq:gan} is set to $\alpha = 0.1$. 

The generated samples are visualized in Figure \ref{fig:mnist} and Figure \ref{fig:celeba}. For MNIST, when a sampler is well trained, the sample distribution over 10 classes on the generated samples should be relatively uniform in order to match that of the training-data statistics. To verify this, we classify the generated samples with a pre-trained deep-CNN MNIST classifier (with a 99.1\% accuracy). We calculate the {\em class distribution} on two settings with different injected noise, $\xib = 0$ and $\xib \sim \mathcal{N}(\mathbf{0}, 0.5\mathbf{I})$. The empirical distributions of the ten digits are indeed even. We also plot the digits generated from the learned Markov chain in Figure \ref{fig:mnist}, from which we can see in the case of $\xib \sim \mathcal{N}(\mathbf{0}, 0.5\mathbf{I})$, the sampler can make distant jumps easily; whereas when $\xib = 0$, transitions seem to be very smooth, thus it needs longer time to generate all digits from the ten classes. Similar results on CelebA, though not as obvious, are observed in Figure~\ref{fig:celeba}. More detailed results are included in the SM.

\section{Conclusion}
Motivated by the WGF theory, we present self-adversarially learned Bayesian sampling, a generative model learning to draw samples from a target distribution. Two settings, {\it i.e.} whether or not true samples are provided as training data, are considered. When learning without true samples, a self-learning mechanism is proposed to automatically adjust samples from the current generator to approach a target distribution. Our method is fully automatic, and is fast and effective in sample generation. Experiments on both synthetic and real datasets demonstrate the effectiveness of our framework, which endows good convergence property while is able to generate much less correlated samples, relative to existing methods. 

\bibliographystyle{aaai}
\bibliography{reference}

\end{document}